\pdfoutput=1

\documentclass[11pt]{article}

\usepackage[]{emnlp2021}

\usepackage{times}
\usepackage{latexsym}
\usepackage{graphicx}
\graphicspath{ {./} }

\usepackage[T1]{fontenc}

\usepackage[utf8]{inputenc}

\usepackage{microtype}

\title{Benchmarks for Indian Legal NLP: A Survey}

\author {Prathamesh Kalamkar\\
  ThoughtWorks \\
  \texttt{prathamk@thoughtworks.com}\\
  Joint first author\\ \AND
  
  Janani Venugopalan \\
  ThoughtWorks \\
  \texttt{janani.venugopalan@thoughtworks.com}\\
  Joint first author\\
  \AND
  Vivek Raghavan\\
  \texttt{vivek@ekstep.org}\\
  \texttt{This work is funded by EkStep}\\
  } 
  
\begin{document}
\addtolength\titlebox{1cm}
\addtolength\titlebox{1cm}
\addtolength\titlebox{1cm}
\maketitle
\begin{abstract}
Availability of challenging benchmarks is the key to advancement of AI in a specific field. Since Legal Text is significantly different than normal English text, there is a need to create separate Natural Language Processing benchmarks for Indian Legal Text which are challenging and focus on tasks specific to Legal Systems.  This will  spur innovation in applications of Natural language Processing for Indian Legal Text and will benefit AI community and Legal fraternity.  We review the existing work in this area and propose ideas to create new benchmarks for Indian Legal Natural Language Processing. 
\end{abstract}
\section{What is an NLP Benchmark}
A machine learning (ML) or artificial intelligence (AI) pipeline typically consists of the following steps: data collection, evaluation and testing, and development of the AI/ ML models. A good solution typically involves the development of high performing models which are evaluated against a pre-specified evaluation criteria. In the past, it has been observed that having specific and challenging unsolved tasks with clearly specified evaluation criteria, has triggered a tremendous spurt of innovation. The exponential growth of community contributed solutions and innovation has been seen to advance the fields of ML such as image processing, natural language processing (NLP), and language models. A benchmark in these fields consists of specific tasks, pre-specified evaluation criteria, and a high-quality data set. 
In NLP, a benchmark is for evaluating the NLP models for a specific task or set of tasks. Some of the famous NLP benchmarks are Stanford Question \& Answer Dataset SQuad2.0 by \cite{rajpurkar2018know} , General Language Understanding  GLUE by \cite{wang2019glue} and super GLUE by \cite{wang2020superglue}, NIST Open MT .  An NLP benchmark would typically  have three main components; training data , testing data and Automatic Model Evaluation system. Training data is used for training the NLP models and made available to all participants. 

Automatic Model Evaluation system is for scoring a given NLP model using testing data which is typically kept hidden. Based on the scores of the NLP models submitted by the ML community, leaderboards are typically created. Machine learning community competes on such benchmarks and this creates a lot of innovative solutions for the problem at hand. Creation of such benchmarks has proven to be the foundation of progress of NLP in an area. Machine Learning Researchers also write research papers describing their solution and many a times open source the code \cite{paperswithcode} for the developed solution. Hence creating a challenging benchmark with the right dataset can spur innovation in desired fields. 

\section{Need For Indian Legal NLP Benchmarks}

While there are some limited official applications of AI and ML techniques in the Indian legal system \cite{OfficialAI}, there exists several avenues and sub-problems where these techniques can be leveraged. Since the legal system generates huge amounts of textual data, NLP techniques can be leveraged to improve the efficiency of repetitive tasks.  To spur innovation in processing of the legal text corpora, we identify some NLP benchmarks which focus on Indian Law systems. They  are needed because of following reasons
\begin{enumerate}
\item  Indian Legal Language is different from general purpose English. Indian Law uses its own language which is esoteric and Latin Language based. Many of the terms used are peculiar and not used in general purpose English. Hence Models pre-trained on general purpose English (like Wikipedia) produce degraded performance on Indian Legal tasks. Hence Indian Legal Benchmark datasets would enable NLP models to learn the legal language.
\item Translation of Relevant Legal tasks to NLP tasks. Application of NLP techniques to solve Legal tasks needs understanding of the NLP world and Legal world. A team of Legal and NLP experts need to brainstorm and come up with relevant legal problems which can be solved with NLP. A good benchmark would be valuable for the legal community and at the same time challenging for the Machine Learning Community. 
\end{enumerate}

\section{How can NLP Benchmarks Help Innovation in Indian Legal System?}
There are currently no NLP benchmarks specifically for Indian Legal system. There are legal NLP benchmarks for other countries \cite{zhong2020does} like China, US and other EU countries \cite{Legal_NLP_Data}. Since India follows Common Law, many of these datasets from countries which follow civil law are not much useful in Indian Context. Also the problems faced by Indian Judiciary are different which are not captured by current legal NLP datasets and problems. Hence defining Indian Legal specific NLP benchmarks along with relevant data will attract the ML community  to solve such unique problems. NLP benchmark will also act as an open evaluation platform for commercial legal NLP solutions providers. Using this platform they can prove the effectiveness of the products and solutions created by them. ML community, startups and researchers will benefit from the knowledge sharing through research papers and open source code thereby helping to solve complex problems in Indian Legal NLP space.

\section{Public Datasets in Indian Law}

To be able to create Indian Legal NLP benchmarks, a lot Indian law specific text data is needed. Using this data , a lot of task specific datasets could be created using human annotations. E.g. For generating Factoid Question and Answer benchmark dataset, humans would annotate the questions based on the court judgement text. Using Indian law text ensures that the NLP models learn the language used in Indian Law. 

\subsection{Data Availability}
Thanks to many open data initiatives like National Judicial Data Grid \cite{NJG} and Crime and Criminal Tracking Network and System \cite{CCTNs}, a lot of data related to law and crime is publicly available.  Table \ref{tab:publicdata} shows the public data sources which can be used for multiple benchmarks.

\begin{table*}
\centering
\begin{tabular}{lll}
\hline
\textbf{Name} & \textbf{Description}\\
\hline
Court Judgements & Districts, High \& Supreme Courts judgements are publicly available on their \\ & websites. These judgements are in English and contain rich text data for Indian Law.\\
FIR by Police & Police Departments of many states provide FIRs on their websites. \\ & Some of them are in local Indian Languages.  \\
India Code & Digital Repository of All central and state acts \\ 
\hline
\end{tabular}
\caption{Publicly Available Sources of Data. Released through Government of India platforms.}
\label{tab:publicdata}
\end{table*}

\subsection{Challenges}
Following are some of the challenges in dealing with the legal documents
\subsubsection{List of judgements not available for all the courts}
The list of published judgements along with a link to the judgement is needed to be able to download the judgement documents. This list is available only for a few high courts. Many high courts give such a list for short periods of time like the past 1 month. Not having such a list limits the amount of information available.

\subsubsection{Inconsistent document formats}
The format of the judgements change with high courts. Some judgments are scanned documents while some are pdfs with text. Scanned documents require additional processing to convert them into machine readable formats.

\subsubsection{Chargesheet which contain the bulk of the case information are not made public}
Although the summary of the case is written in the court judgement, most of the relevant information for the case is present in chargesheets. This limits benchmarks that can be created with such limited information. E.g. Timeline Summary creation needs events to be described by time in the text. Such information is not present in typical judgement. 

\subsubsection{FIRs are not always updated and often contain a mixture of languages.}
FIRs published by police provide rich but limited information about the case. Publicly available FIRs do not have updated information after further investigation are done.

\section{Indian Legal NLP Benchmark Areas}

\begin{figure*}[htp]
    \centering
    \includegraphics[scale=0.35]{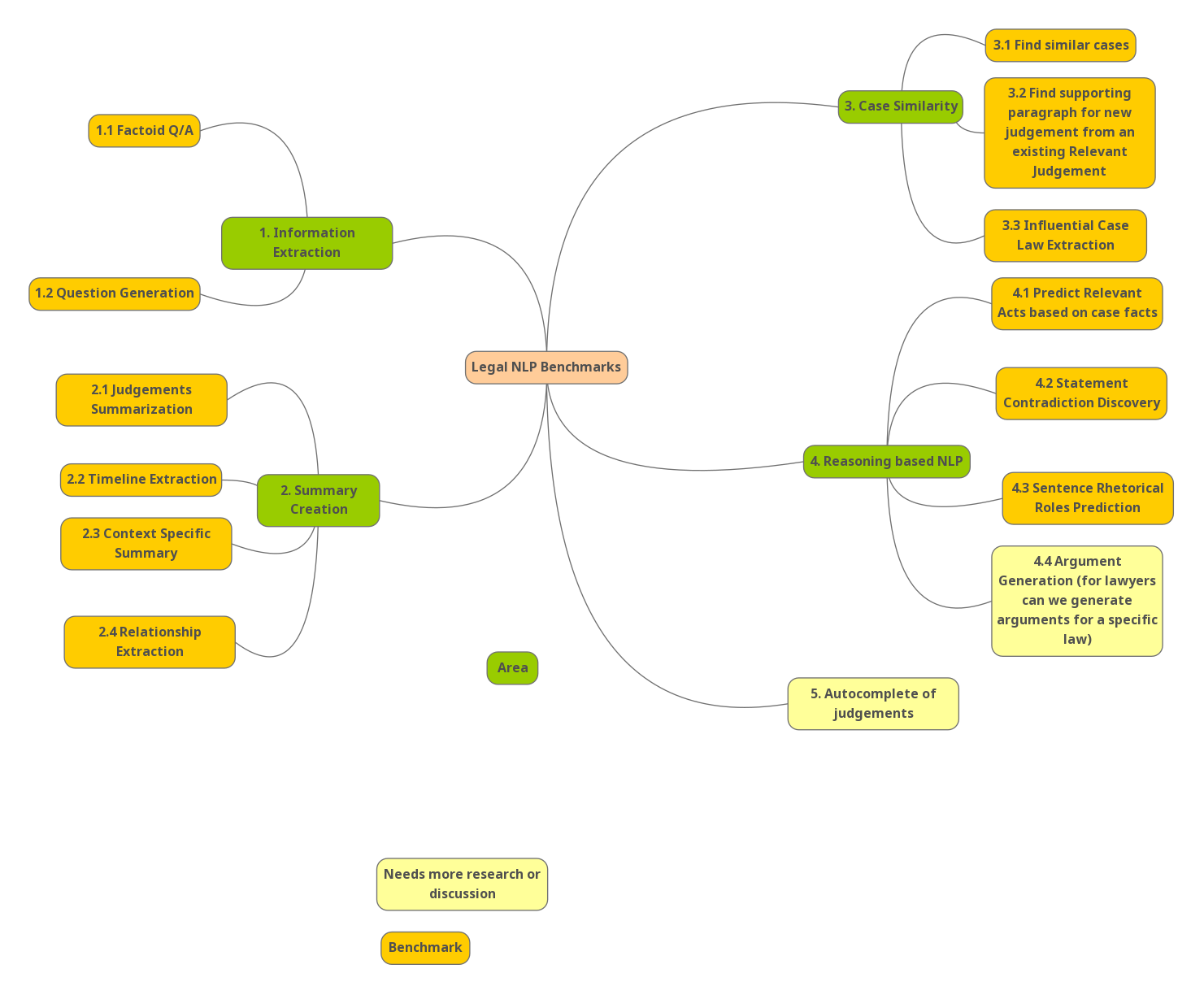}
    \caption{Indian Legal Bench Marks: An Overview}
    \label{fig:benchmark_overview}
\end{figure*}

Mind map Figure \ref{fig:benchmark_overview} shows the NLP areas and NLP benchmarks in each of the areas. 4 main NLP areas have been identified and are marked in green color in the mind map below. 

\subsection{Information Extraction}
Information Extraction in the area of Natural Language Processing deals with extracting the information of interest from text. Following benchmarks identify specific tasks within Information Retrieval which are useful for Indian Legal tasks.

\subsubsection{Factoid Question \& Answers}
\textbf{
\begin{itemize}
\item Description of Task
\end{itemize}
}
Judges and lawyers are looking for certain information based on the type of case. This can be framed as some template questions that need to be answered from the case text. Along with the template questions, there would be some context specific questions.. E.g. For a criminal case, template questions could be “What was the place of occurrence?” , “What was the time of occurrence?” , “Is there any eye-witness?” , “What are Number of Heads and Sections under which charges were framed?” etc. If the case is murder case then along with the template questions the context specific questions could be “What was the murder weapon?”, “What is Nature of injuries mentioned in the Post-mortem Report?” etc. 
So this can be treated as factoid questions i.e. you are looking for answers which are written explicitly in the text as a span of few contiguous words. Factoid Q/A system will find such answers written in text and highlight the answers in the text. E.g. Question of interest is “What is the place of occurence?” The Q/A system searches all the documents and finds relevant passages that have answers to this question and highlight it like in the text below.
“… e of prosecution alleged incident took place at the tea stall situated near  Bombay Central Bus Stand at around 7.30 a.m. and as such there is every…….”
\textbf{
\begin{itemize}
\item Value Proposition
\end{itemize}
}
This automatic extraction of answers to such questions would significantly save the human reading time. Having a challenging benchmark for Factoid Q/A would attract ML practitioners to focus on this problem. This will create a foundation for innovative approaches and open source models focussing specifically on Indian Legal factoid Q/A. 
\textbf{
\begin{itemize}
\item Mapping to standard NLP Tasks
\end{itemize}
}
This benchmark maps to a NLP task called factoid Q/A. This is a widely studied task in NLP and many pretrained NLP models are available which focus on general purpose English language. Given availability of data and many published approaches, this NLP task falls into the easy category.
\textbf{
\begin{itemize}
\item Similar NLP benchmarks
\end{itemize}
}
Similar benchmarks exist for open domain tasks like general wikipedia factoid Question \& answers (Squad2.0 \cite{rajpurkar2016squad}, \cite{rajpurkar2018know}, ComQA \cite{abujabal-etal-2019-comqa}, Google Natural Questions \cite{kwiatkowski2019natural} etc.) But these datasets lack the domain specific legal language which limits use of such benchmarks. There is also a reasoning based legal dataset (JEC-QA \cite{zhong2020jec}) which involves complex questions based on the questions from the National Judicial Examination of China. These questions need prior knowledge of law and reasoning. Hence such datasets from civil law countries and using general English text are not relevant in the context of Indian Legal domain. Hence a factoid question and answer dataset using Indian Legal text would ensure that the solutions developed work well with Indian Legal Language.
\textbf{
\begin{itemize}
\item Dataset to be collected
\end{itemize}
}
Data needed for this benchmark would be template questions about a given legal text and humans would mark the answers in that text. The person will also create the context specific questions which could be generated by him/her or taken from the automatically generated context specific questions (as mentioned in the next question generation benchmark). For creation of such human annotated data , existing tools like haystack \cite{haystackAI} or cdQA \cite{CDQA} could be used with some changes needed. The human annotations could be done by a literate person with basic reading comprehension skills and some training of the legal language. 
\textbf{
\begin{itemize}
\item Evaluation Metrics
\end{itemize}
}
The evaluation of this benchmark would be done by matching the answers generated by NLP models with human extracted answers. Evaluation Metric could be F1 score using exact match.

\subsubsection{Question Generation}
\begin{figure*}[htp]
    \centering
    \includegraphics[scale=1]{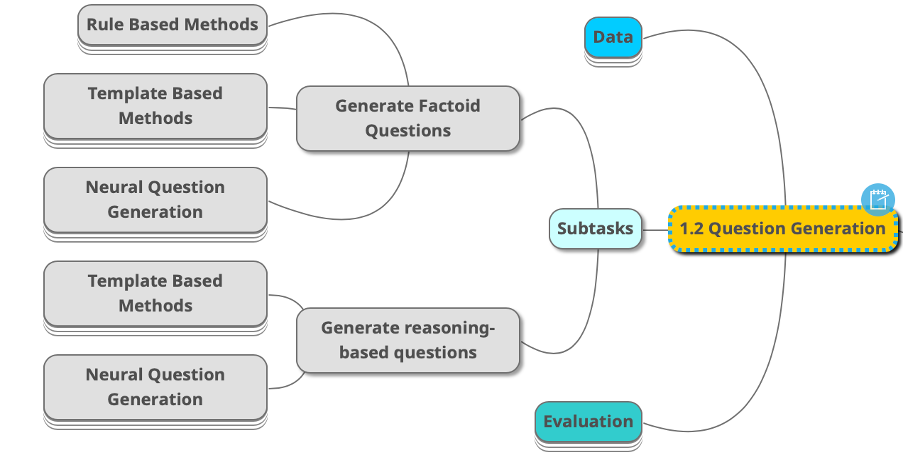}
    \caption{Question generation using state of the art techniques.}
    \label{fig:question generation}
\end{figure*}

\textbf{
\begin{itemize}
\item Description of Task
\end{itemize}
}
In order to get meaningful insights from answers of questions, it is important to ask the right questions which are context specific. These  questions can be of two types: factoid and non-factoid or reasoning based questions. Factoid questions are the questions for which the answers are explicitly stated in text as facts of contiguous text. E.g. Context: "… companion were thieves and therefore took away driving licence of Mohammad Nisar Khan along with some visiting cards and coins. At the same time took ..."  From this context the generated factoid Question would be  “What is the name of the victim?” 
Reasoning based questions are those for which the answers have to be created using information written in text along with using some logic. E.g. Context: "… companion were thieves and therefore took away the driving licence of Mohammad Nisar Khan along with some visiting cards and coins. At the same time took ...". Based on this context, a reasoning based question would be , “Did Mohammad Nissar Khan contradict himself when he gave evidence between the district and the high court?” 
Generation of reasoning based questions and answering is more difficult than factoid based questions. Hence it would be useful to start with factoid based questions and then move to reasoning based questions. 
Recent advancements in Neural question generation allow us to create answer-aware questions i.e. given a context and an answer , the question can be generated. So one can create a lot of questions from a context using different answers. Hence it is important to identify which information in the context is important and generate correct questions which would lead to those answers. Hence the candidate model should submit the question answer pairs for a given judgement. These QA pairs would be compared with the Human generated QA pairs to evaluate how good the question generation is. 
\textbf{
\begin{itemize}
\item Value Proposition
\end{itemize}
}
For each case it is very imperative that the right questions are asked. Sometimes, even the outcome of the case is dependent on the type of questions asked. This question generation is a very time-consuming process and often needs the documents to be perused several times. In this task, we propose the use of NLP to automate the bulk of question generation, allowing lawyers and others in the legal system to focus on the game-changing questions. 
In addition, context specific question generation also helps in collecting human annotated data for the creation of the NLP benchmarks which can be by the technology community In many benchmark datasets like Squad \cite{rajpurkar2016squad}, \cite{rajpurkar2018know}, the questions are generated by humans by looking at the text and answers are marked in that text. A lot of time would be saved if the questions are also generated automatically and humans just have to mark the answers in the text. 
\textbf{
\begin{itemize}
\item Mapping to standard NLP Tasks
\end{itemize}
}
This benchmark maps to a NLP task called question generation and language generation.With recent advancements in question answering, question generation, there are several studies which focus on both factoid and reasoning based question generation. The NLP models for this are also focused on general purpose English language. Given availability of data and some published approaches, this NLP task falls into the moderate category. The different approaches can be summarized in Figure \ref{fig:question generation}
\textbf{
\begin{itemize}
\item Similar NLP benchmarks
\end{itemize}
}
Similar benchmarks exist for selection specific question generation benchmarks such as Sel-QA \cite{7814688}, Wiki-QA \cite{yang-etal-2015-wikiqa}, and GlGE \cite{liu2020glge}. They are taken from a curated list of topics from corpora such as wikipedia. They are essentially question-answering benchmarks where the questions are generated based on a selected text, and cover various versions of the questions through human annotations. The questions generated are essentially evaluated for coverage of the text and linguistic correctness. But these datasets lack the domain specific legal language which limits use of such benchmarks. There is also a reasoning based legal dataset (JEC-QA \cite{zhong2020jec}) which involves complex questions based on the questions from the National Judicial Examination of China. However, this is a question answering benchmark and not a question generation one. Hence a generation dataset using Indian Legal text would ensure that the solutions developed work well with Indian Legal Language and are specific to this task.
\textbf{
\begin{itemize}
\item Dataset to be collected
\end{itemize}
}
The data for this task could be collected from judgments, FIRs, chargesheets. Of these sources, the judgments from district-courts, high-courts, and the supreme-courts of India are available for use in such tasks. The humans will mark the QA pairs generated by a baseline model as valid, and also generate additional QA pairs for the same selected piece of text to get as many variations as possible. This is similar to methods employed in other question generation benchmarks. We need only people who are proficient in the language, not necessarily legal experts, for creating the data in this benchmark. We can leverage existing tools such as  ProphetNET \cite{qi2021prophetnet}, T5 \cite{kumar2021deep} and Squash \cite{krishna2019generating} fine-tune them if needed for Legal question generation.
\textbf{
\begin{itemize}
\item Evaluation Metrics
\end{itemize}
}
There are two main aspects of the evaluation of this benchmark. First one is whether the QG process has identified the right answers from context to create questions (called as coverage). The second is to check the quality of question text. For coverage of the answers, the generated answers would be compared with Human answers to calculate the overlap. The Quality of the questions could be checked by checking the grammar of the questions, passing the question through the standard question answering model to check whether the answer matches with the given answer.
\subsection{Summary Generation}
\subsubsection{Judgement Summarization}
\textbf{
\begin{itemize}
\item Description of Task
\end{itemize}
}
The court judgements, especially from high courts and supreme court, tend to be very long. Finding the right context from such long judgement is time consuming and error prone. Supreme Court judgements from 1950 to 1994 used to have summaries created by lawyers while publishing. But after that the summaries are not present in the supreme court judgement texts. This benchmark would focus on evaluating various aspects of the summaries created for a given judgement. 
\textbf{
\begin{itemize}
\item Value Proposition
\end{itemize}
}
The summary of high courts and supreme court is often used in establishing precedent. Searching in summaries rather than entire text would help lawyers to establish better arguments for precedence. Summary of judgements from lower courts also help in reducing processing time when case moves to higher court.
\textbf{
\begin{itemize}
\item Mapping to standard NLP Tasks
\end{itemize}
}
Summarization of texts is a standard NLP task. There are two types of summaries; extractive and abstractive. Extractive summaries focus on extracting important complete sentences from text to form the summary. Abstractive summaries on the other hand create summaries by constructing new sentences which effectively summarizes the text. 
\textbf{
\begin{itemize}
\item Similar NLP benchmarks
\end{itemize}
}
While there are no summarization benchmarks that focus on the legal documents, there are many benchmarks that focus on summarizing general English text like Wikipedia articles, news articles \cite{narayan2018don}, CNN/dailynews \cite{nallapati2016abstractive}, research(ArXiv) articles \cite{leskovec2005graphs}, \cite{gehrke2003overview}, pubmed articles \cite{Sen_Namata_Bilgic_Getoor_Galligher_Eliassi-Rad_2008}.  Similarly, the current state of the art research focuses on comparing the performance of various algorithms on Indian Judgements \cite{bhattacharya2019comparative}. 
\textbf{
\begin{itemize}
\item Dataset to be collected
\end{itemize}
}
Since the summaries created by lawyers of the supreme court for judgments from 1950 to 1994 are available in the form of headnotes,  this data could be utilized to create the benchmark.
\textbf{
\begin{itemize}
\item Evaluation Metrics
\end{itemize}
}
The automatic evaluation of summaries created is a complex topic.  Recent research \cite{fabbri2021summeval} proposes that there are 4 main aspects of summary evaluation: Coherence, Consistency , Fluency \& Relevance. Many of the existing evaluation metrics like ROUGE, METEOR etc. depend on availability of the human generated summary to compare against. But these metrics do not capture the Faithfulness and factuality \cite{maynez2020faithfulness} of the generated summaries. The proposed ways to capture Hallucinations are computationally expensive. Hence there is a need to formulate better evaluation metrics which can measure faithfulness and factuality efficiently. 

\subsubsection{Timeline Extraction}
\begin{figure*}[htp]
    \centering
    \includegraphics[scale=0.35]{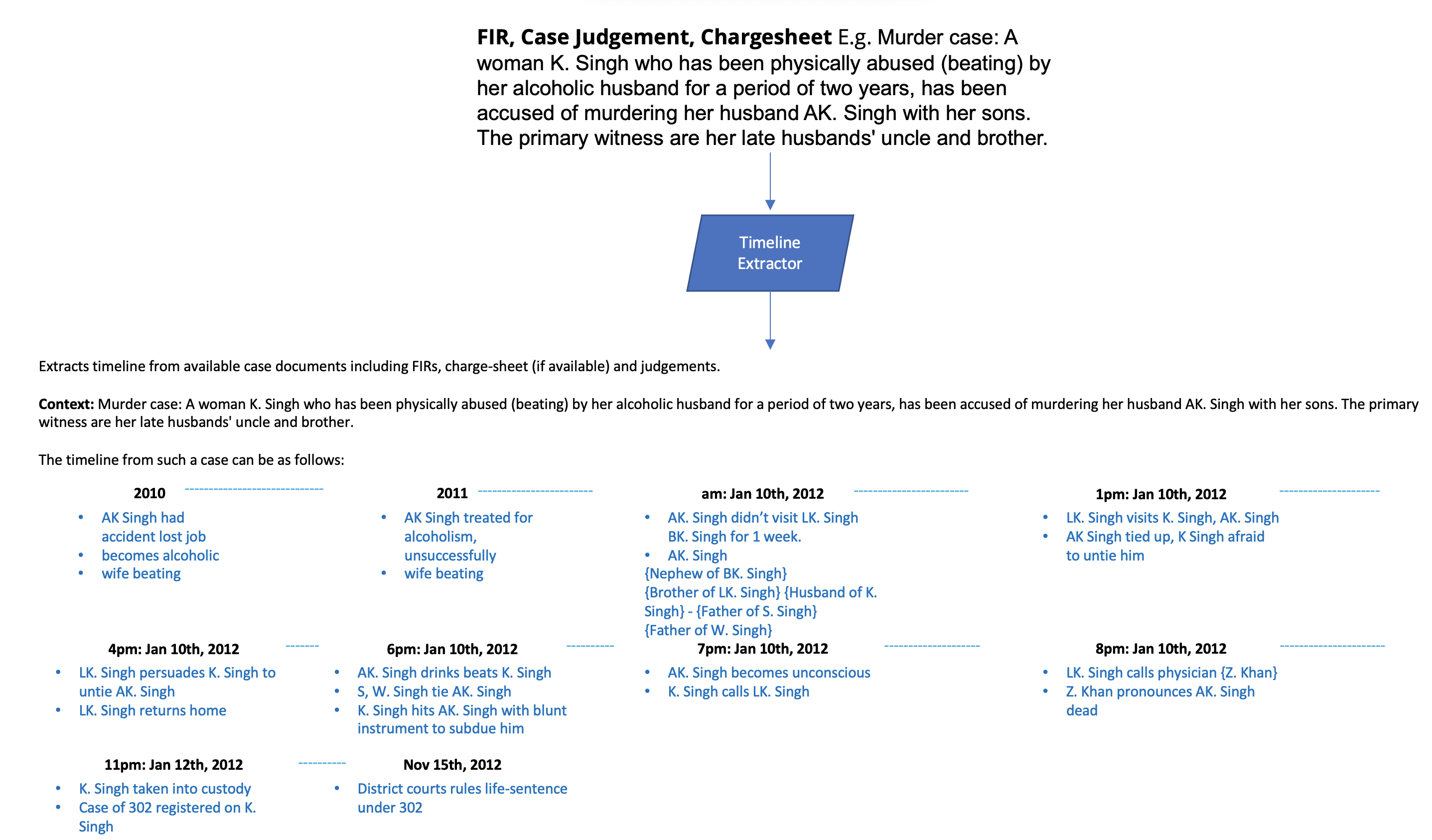}
    \caption{Context timeline extractor system which takes the case text (e.g. judgement, FIRs, chargesheet etc.) to get a timeline of events.}
    \label{fig:timeline summary example}
\end{figure*}

\begin{figure*}[htp]
    \centering
    \includegraphics[scale=0.4]{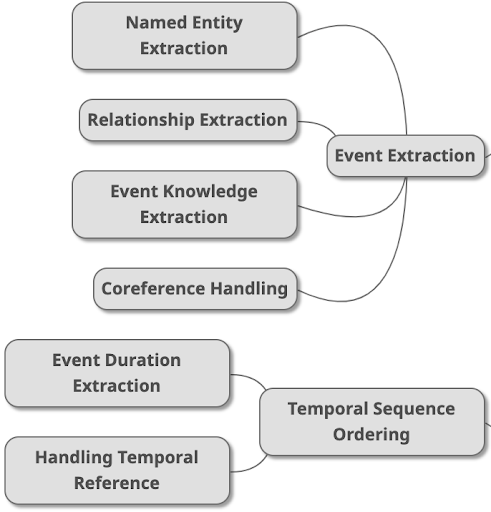}
    \caption{Timeline generation using state of the art techniques.}
    \label{fig:timeline generation}
\end{figure*}

\textbf{
\begin{itemize}
\item Description of Task
\end{itemize}
}
In this use-case scenario we propose the use of NLP to extract timelines from these case documents (Figure \ref{fig:timeline summary example}).  In this white paper, we propose a timeline extraction benchmark to be of use to both the legal informatics community and the machine learning (ML) community. 
\textbf{
\begin{itemize}
\item Value Proposition
\end{itemize}
}
Typically a judgment in the Supreme Court of India  is several pages long and has information pertaining to different aspects of the case including  previous judgements, witness accounts, references, arguments and counterarguments. In addition to these, the cases may have one or more of the following; chargesheet, FIRs, lower court judgements and orders, district, high court and appellate court judgements and orders. As a result, depending on the type of case, each case may have hundreds or even thousands of pages worth documentation. The legal representation and decision making process for each court case is also handled by multiple people. 
The fact that the cases are often handled by multiple lawyers and judges, extracting the facts of the case accordings to events which occurred in time can get very challenging. With witness statements, and arguments being recorded several times, this can get even more challenging.
\textbf{
\begin{itemize}
\item Mapping to standard NLP Tasks
\end{itemize}
}
This benchmark maps to a NLP task called temporal event extraction. There is a very nascent and growing field in NLP with a few seminal studies. The NLP models for this are also focused on general purpose English language. The different approaches can be summarized in Figure \ref{fig:timeline generation}.
To create targeted models which go on the leaderboard for this benchmark, one can leverage existing timeline extraction techniques including event extraction and temporal sequence techniques \cite{piskorski2020timelines},  \cite{ning2018cogcomptime}, \cite{chieu2004query}, \cite{finlaysonaextracting}, \cite{yu2020spatiotemporal} Figure \ref{fig:timeline generation}.
Given availability of data and some published approaches, this NLP task falls into the moderate- difficult category.
\textbf{
\begin{itemize}
\item Similar NLP benchmarks
\end{itemize}
}
Despite the existence of a few companies \cite{EventRegistry}, \cite{PrimerAI}, there exist very few open event, timeline extraction datasets and benchmarks in the ML community e.g.[CoNLL2012 \cite{pradhan2012conll}, SemEval2015 \cite{minard2015semeval}]. A lot of works rely on text such as WikiPedia and news articles.  A major challenge with this is that there exist very few temporal relationships in these datasets or the studies only extract dated events. In this whitepaper, we propose a timeline extraction benchmark to fill these gaps and also to provide a brand new legal timeline extraction benchmark. 
\textbf{
\begin{itemize}
\item Dataset to be collected
\end{itemize}
}
The data for this task could be collected from judgements, FIRs, chargesheets. Of these sources, the judgements from district-courts, high-courts, and the supreme-courts of India are available for use in such tasks. The training/ test data we propose are human extracted events ordered in time. We need only people who are proficient in the language, not necessarily legal experts, for creating the data in this benchmark. To assist in the creation of the training and test data, we will create an initial baseline model for extracting timelines and events from a selected text. The annotator has to add missing information and reorder timelines where necessary. 
\textbf{
\begin{itemize}
\item Evaluation Metrics
\end{itemize}
}
The automatically generated benchmarks are evaluated against the expert created ones using accuracy and degree of overlap metrics.

\subsubsection{Context specific Summary}
\begin{figure*}[htp]
    \centering
    \includegraphics[scale=0.6]{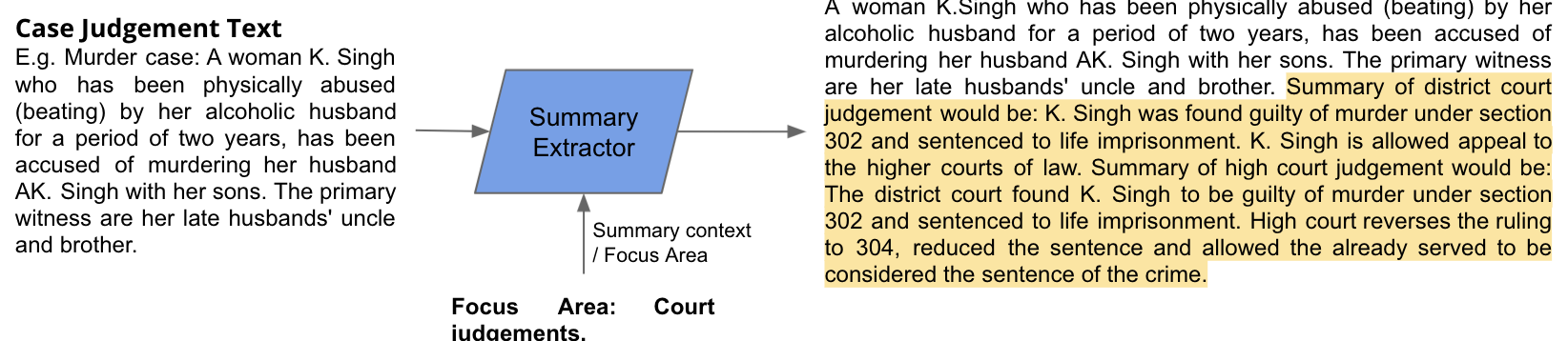}
    \caption{Context specific summary system takes the case text (e.g. judgement, FIRs, chargesheet etc.) and the focus area (e.g. court judgements) to get a summary specific to the focus area}
    \label{fig:context specific summary}
\end{figure*}

\textbf{
\begin{itemize}
\item Description of Task
\end{itemize}
}
Context specific summary systems take the case text (e.g. judgement, FIRs, chargesheet etc.) and the focus area (e.g. court judgements) to get a summary specific to the given focus area. An example of a context specific summary is shown in figure \ref{fig:context specific summary}.
\textbf{
\begin{itemize}
\item Value Proposition
\end{itemize}
}
As mentioned above, the case documents can run to a few hundred or even thousands of pages, with multiple legal representation at each stage. It will be very useful to the legal community, if they had access to context specific summaries e.g. summary of the statements given by primary witnesses across the different courts. It will be a very time-saving component for the legal system.
\textbf{
\begin{itemize}
\item Mapping to standard NLP Tasks
\end{itemize}
}
This benchmark maps to summarization and as an extension, entity specific summarization in NLP. Currently, the state of the art ML summarization focuses on single \cite{haque2013literature} and multi-document summaries \cite{batra2020review} for the entire text. The different types of solutions include,  concept map based benchmarks \cite{falke2017bringing}, machine-generated and human generated text based solutions. Similarly, the state of the art legal benchmarks contain summary generation tools trained on legal documents \cite{farzindar2004legal}, \cite{kanapala2019text}, \cite{farzindar2004letsum} including the Indian Legal Documents \cite{8554831}, \cite{bhattacharya2019comparative}, which they tested using human generated summaries \cite{WestLaw}. In addition to these there are text summarization tracks in NLP conferences such as Text REtrieval Conference \cite{TextAnalysis}, Text Analysis Conference \cite{TextAnalysis}, and Forum for Information Retrieval Evaluation \cite{FIRE}.  
\begin{figure*}[htp]
    \centering
    \includegraphics[scale=0.7]{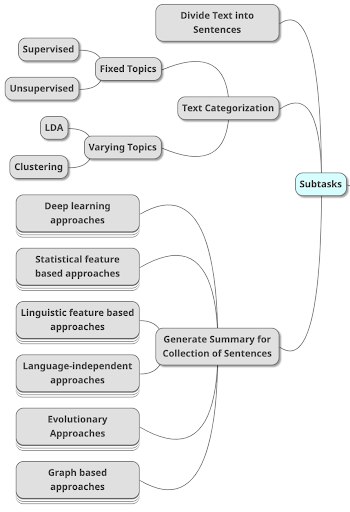}
    \caption{Context specific summary generation using current summarization techniques}
    \label{fig:context specific summary process}
\end{figure*}
To create targeted models which go on the leaderboard for this benchmark, one can leverage existing summary generation models such as context specific sentence extraction and text categorization and subsequent summary generation methods. Some examples of the said methods are shown in figure \ref{fig:context specific summary process}.
\textbf{
\begin{itemize}
\item Similar NLP benchmarks
\end{itemize}
}
In this white paper, we propose a context specific summarization benchmark to be of use to both the legal informatics community and the machine learning (ML) community. However, to the best of our knowledge, there exists no ML and legal benchmarks on context/ actor specific summary generation. 
\begin{figure*}[htp]
    \centering
    \includegraphics[scale=0.7]{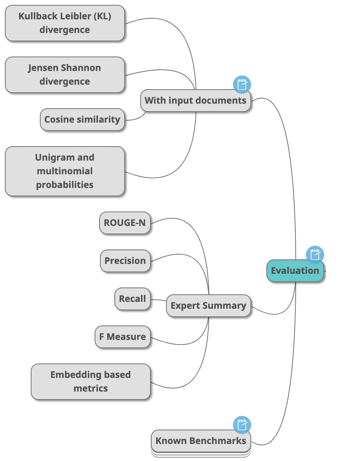}
    \caption{Context specific summary evaluation: Metrics can be applied either on the input documents or expert generated summaries. Figure shows the evaluation metrics which can be applied to either of the two scenarios}
    \label{fig:context specific summary evaluation}
\end{figure*}
\textbf{
\begin{itemize}
\item Dataset to be collected
\end{itemize}
}
In our benchmark, we propose the use of open documents from the Indian Legal System (e.g. case judgements) as the dataset. The training/ test data here can be either the full-text and the human labelled summaries. The focus areas for this benchmark will be limited to the actors identified in each case using named entity recognition (relationship extraction)  and a few expert curated focus areas such as judgments, citations, primary witness statements, introduction, context, conclusion, arguments, counter-arguments etc. If we use expert created summaries , then we need legal experts to create the summaries. However, it is perfectly valid in this benchmark to not use expert summaries but validate against input data.
\textbf{
\begin{itemize}
\item Evaluation Metrics
\end{itemize}
}
The evaluation metrics (Figure below) can be done with respect to the input documents or with expert generated benchmarks. When we compare with the input documents, we can rely on evaluation metrics which can get the similarity between sequences of different lengths (e.g. Kullback Liebler divergence, Jensen Shannon divergence, cosine similarity). When we use expert generated summaries for the evaluation, we can rely on the metrics which also account for the summary lengths in addition to similarity (e.g. precision, ROUGE-N, Recall, F measure, embedding based metrics).

\subsubsection{Relationship Extraction}
\textbf{
\begin{itemize}
\item Description of Task
\end{itemize}
}
The legal text describes multiple relations between the entities involved. These entities could be person , place , acts , court , judges etc. The extracted textual relation can be categorized into predefined relation categories. This allows us to filter for interesting relations. 
The extracted relations can be visualized in the form of a graph. Indirect relations can be inferred using this graph. 
E.g. Case Judgement Text says ,“Complainer: Ashok Sethi ... Accused: Pranav Jog … ….Compainer \& accused live in the same society...Therefore, accused assaulted complainer with rod....”. Following relations could be found out from this text.
\begin{table*}
\centering
\begin{tabular}{lll}
\hline
\textbf{Entity1} & \textbf{Relation} & {\textbf{Entity2}}\\
\hline
Ashok Sethi & is & Complainer\\
 Pranav Jog & is & Accussed \\
 Complainer & is neighbor & Accussed\\
 Complainer & assaulted & Accussed \\
\hline
\end{tabular}
\caption{Example of Extracted Relations}
\label{tab:extractedrelations}
\end{table*}
Relationship extraction can be treated as an unsupervised task where the entities , relations are text. But categorization of extracted relationship text into meaningful categories needs machine learning model training. E.g. mapping of “live in same society” to “is neighbor”.
\textbf{
\begin{itemize}
\item Value Proposition
\end{itemize}
}
This task would provide a quick Summary of relation between entities in text. Indirect relations can easily be inferred using graphs. The relations would be ones that are important from a legal perspective. Having a challenging benchmark for Indian Legal Relation Extraction would attract ML practitioners to focus on this problem. This will create a foundation for innovative approaches and open source models focussing specifically on Indian Legal Relation Extraction. 
\textbf{
\begin{itemize}
\item Mapping to standard NLP Tasks
\end{itemize}
}
This benchmark maps to Knowledge Base Population \cite{KBP} , Open Information Extraction\cite{OpenIE}. Both of the NLP tasks are harder to evaluate and hence tricky to improve on. Most of the research focuses on open domain text and using ontologies to get domain specific relations.  Many of the existing NLP components like Legal Named Entity Relation Extractor \cite{Blackstone} , open information extraction \cite{OpenIE} can be leveraged for this. 
\textbf{
\begin{itemize}
\item Similar NLP benchmarks
\end{itemize}
}
There are many existing benchmarks which deal with open domain relation extraction from text documents like wikipedia or news articles (\cite{yao2019docred}, \cite{zhang2017tacred}, \cite{10.1007/978-3-642-15939-8_10} etc.). But none of them focus on relations that are interesting in legal documents like a person is a complainer or accused, a person is a judge etc. Creating Indian legal domain specific relations and benchmarks would trigger innovation in solutions that perform better on relations in Indian Legal Text . 
\textbf{
\begin{itemize}
\item Dataset to be collected
\end{itemize}
}
The data needed for this benchmark is Human extracted relations in given text. Humans would be presented with text and categorized relations. Humans would then extract the relations between the entities as written in text and map them to the categorized relations. Skill Sets needed for human annotations would be understanding the basics of Legal language and relations. These could be easily picked by a layman literate person with basic training. A custom annotator tool needs to be built to allow such human annotations. 
\textbf{
\begin{itemize}
\item Evaluation Metrics
\end{itemize}
}
The evaluation of this benchmark could be done based on matching the relations extracted by candidate NLP model with the human extracted relations. NLP Models should be able to correctly identify most of the relations and correctly categorize those.

\subsection{Case Similarity}
Indian Law is common law. This means that precedent set by previous courts to interpret laws must be respected. Hence finding out cases which are similar to a given case becomes important for legal research. This activity is manual and a lot of commercial tools use keyword based or act based similarities. But to be able to search the similar judgements based on longer text like case facts, FIR etc. would make the search more relevant and automated. Following benchmarks focus on these tasks.

\subsubsection{Find Similar Cases}
\textbf{
\begin{itemize}
\item Description of Task
\end{itemize}
}
This benchmark deals with finding most relevant cases for given text descriptions about a case. Many existing search engines of Indian Legal documents (India Kanoon , Legitquest, aironline etc.) focus on searches based on keywords. Searching with text descriptions allows you to look for more detailed information rather than keywords based search. Hence one can input the FIR, chargesheets , case description and search for the judgements which are similar.  E.g. Case description says ,”Republic Editor Arnab Goswami was arrested in criminal case of suicide. His claims that he was targeted by the state govt and his personal liberty was violated”. Then similar cases might be judgements that interpret personal liberty like  Maneka Gandhi v. Union of India and Another (1978) , Kharak Singh v. State of U.P. and Others etc. The similar cases could come from other high courts or supreme courts. 
\textbf{
\begin{itemize}
\item Value Proposition
\end{itemize}
}
Reduce time needed for manual search of similar cases and provide more meaningful results. Provision of search by larger text instead of just keywords would be very helpful. 
Many startups are developing such search engines as well. But there is no benchmark about how good the search engines are. Hence this benchmark would provide an objective way of evaluating such products
\textbf{
\begin{itemize}
\item Mapping to standard NLP Tasks
\end{itemize}
}
This maps to case2vec where the cases are mapped to vectors in latent space. This mapping of cases to latent vectors is learnt while training. These types of tasks are harder to evaluate and hence hard to improve on.
\textbf{
\begin{itemize}
\item Similar NLP benchmarks
\end{itemize}
}
In the Forum of Information Retrieval Evaluation 2019 \cite{inproceedings}, there is a dataset created which identifies the most similar cases for a given text description of a case from a set of 2914 supreme court judgements. This data is available for 50 text description queries. 
Chinese AI Law competition \cite{xiao2019cail2019scm} has created a similar case matching competition (CAIL 2019 - SCM). This competition focuses on finding which 2 supreme court cases are similar in a triplet of cases. But the data is in Chinese Language. Other Existing benchmarks focus on finding similarity between sentences (Semantic Textual Similarity Benchmark and Microsoft Research Paraphrase Corpus). But these focus on open domain sentences similarity and not on similarity of the entire document. Competition on Legal Information Extraction/Entailment \cite{COLIEE2020} focuses on extracting supporting cases for a given new case using Canada cases. 
 Hence there is a need for benchmark of Indian Legal cases similarity.   
\textbf{
\begin{itemize}
\item Dataset to be collected
\end{itemize}
}
The data needed for this benchmark is past judgements and metadata about the case like acts etc. Using this data, triplets of cases (A,B,C) would be made. Legal Experts would tag if case A is more similar to B or C. Such triplets need to be created carefully so that cases in a triplet share some common factors and are not random.This is because it is much easier to distinguish between two completely different cases like land case vs. murder case than to find similarity of one murder case to other 2 murder cases. A custom annotator tool needs to be built to allow such human annotations. 
Legal experts are needed for the manual tagging of such triplets and multiple such opinions would be taken for a record. Consensus of experts would be used as final data. 
\textbf{
\begin{itemize}
\item Evaluation Metrics
\end{itemize}
}
This task can be evaluated by comparing model predictions with the consensus of the legal experts. The score could be the accuracy of the predictions. 

\subsubsection{Find supporting paragraph for new judgement from an existing Relevant Judgement}
\textbf{
\begin{itemize}
\item Description of Task
\end{itemize}
}
Many times while writing judgements , a judge would take reference of an old judgement which has interpreted law in detail. This is a crucial step to make sure case precedent is followed. A Judge may already know such existing relevant judgements or use the search system explained in the previous benchmark. Since the judgment is similar, there could be multiple paragraphs that support the new decision. So finding the exact paragraph which supports the new judgement can be time consuming. 
E.g. A review case of Visa rejection comes to a judge where the appellant says that the visa was rejected without an interview by the Visa officer based on information collected by other people. Judge is writing a judgement where he wants to write “in matters of administrative decisions, the rule of “he who hears must decide” does not apply” . Judge has found an existing relevant judgement based on similarity search. Now the judge wants the exact paragraph from the existing judgement which interprets this. So he searches using query as “in matters of administrative decisions, the rule of “he who hears must decide” does not apply” and gives an existing document. The system returns the paragraph from existing judgement which interpretes this law in detail. “The decision is essentially an administrative one, made in the exercise of discretion by the visa officer. There is no requirement in the circumstances of this or any other case that he personally interview a visa applicant. There may be circumstances where failure to do so could constitute unfairness, but I am not persuaded that is the case here. Here the IPO did interview the applicant and reported on the results of that interview. That report was considered by the visa officer who made the decision. Staff processing and reporting on applications is a normal part of many administrative processes and it is not surprising it was here that followed. This is not a circumstance of a judicial or quasi judicial decision by the visa officer which would attract the principle that he who hears must decide, or the reverse that he who decides must hear the applicant.”
\textbf{
\begin{itemize}
\item Value Proposition
\end{itemize}
}
Time saved to find supporting exact text from existing judgement would enable legal stakeholders to process the decisions faster. 
\textbf{
\begin{itemize}
\item Mapping to standard NLP Tasks
\end{itemize}
}
This maps to paragraph2vec where similarity between new judgment text and paragraphs from a given judgement are found out. These type of tasks are hard to evaluate and hence hard to improve on.
\textbf{
\begin{itemize}
\item Similar NLP benchmarks
\end{itemize}
}
There is an existing similar benchmark by Competition on Legal Information Extraction/Entailment (\cite{COLIEE2020}) which focuses on extracting an entailing paragraph from a relevant case for a new judgement using Canada cases. Creating a similar benchmark for Indian Law would be more useful. 
\textbf{
\begin{itemize}
\item Dataset to be collected
\end{itemize}
}
Data needed for this benchmark would be created using Indian Courts judgements. Legal experts would create a triplet (relevant case text , new judgement, paragraph id supporting new judgement). Evaluation of this benchmark would be done by matching the paragraph extracted by humans to paragraphs extracted by model. A custom annotator tool needs to be built to allow such human annotations. The annotator would show the new judgement line and suggest paragraphs based on NLP model output. The answer suggested can be accepted by the expert or he can change it to an appropriate one. This suggestion would reduce the human processing time significantly.
\textbf{
\begin{itemize}
\item Evaluation Metrics
\end{itemize}
}
The evaluation of this benchmark can be done by comparing if the paragraph id predicted by the model matches the one provided by the expert. F1 score of the match can be used to rank the submissions.

\subsubsection{Influential Case Law Extraction}
\textbf{
\begin{itemize}
\item Description of Task
\end{itemize}
}
Some of the judgements have far reaching implications and are commonly cited in multiple judgements. The idea of this benchmark is to objectively identify such influential judgements for each of the Legal Topics. Some of such cases are mentioned in the table \ref{tab:influentialjudgements}.
\begin{table*}
\centering
\begin{tabular}{lll}
\hline
\textbf{Legal Topic} & \textbf{Influential Judgement} \\
\hline
Right to personal liberty & Maneka Gandhi Vs. Union of India , 1978 , ...\\
 Supreme Court’s authority over the\\
 Constitution & Kesavananda Bharati vs state of Kerala , 1973,..... \\
 Uniform Civil Code & Mohammed Ahmed Khan v. Shah Bano Begum , 1985,...\\
\hline
\end{tabular}
\caption{Examples of Influential Judgements}
\label{tab:influentialjudgements}
\end{table*}
Many supreme court judgements cite other judgements for interpretation of laws. Such citations can be extracted automatically with NLP techniques and a network of such citations can be created. Using such a network, one can find influential judgements overall and for specific legal topics. 
\textbf{
\begin{itemize}
\item Value Proposition
\end{itemize}
}
The benchmark would provide objective evaluation of influence of a judgement. This would also mean time saving for legal research.
\textbf{
\begin{itemize}
\item Mapping to standard NLP Tasks
\end{itemize}
}
This maps to citations extraction and algorithms like pagerank to decide influence once the network of citations is created. These NLP tasks are considered easy because of availability of pretrained models and libraries.
\textbf{
\begin{itemize}
\item Similar NLP benchmarks
\end{itemize}
}
There is an existing benchmark that focuses on prediction of case importance for European Human Rights Cases \cite{chalkidis2019neural}.  To the best of our knowledge, there are no benchmarks that focus on establishing influential judgments in specific areas of Indian Law.
\textbf{
\begin{itemize}
\item Dataset to be collected
\end{itemize}
}
Data needed for such benchmark is supreme and high court judgements with citations. Legal experts would be needed to create such lists. Opinions of multiple legal experts need to be combined to create the curated list in specific areas. Since data to be collected involves ranked lists in a given area, complex tools may not be needed for data collection. Experts can use simple tools like MS Excel to create such lists. 
Many of the existing NLP components like Legal Named Entity Relation Extractor can be leveraged for creating structured data for human annotations. 
\textbf{
\begin{itemize}
\item Evaluation Metrics
\end{itemize}
}
Evaluation of this benchmark could be done in a similar way like case similarity benchmark above by matching the model created ranked list of expert created list

\subsection{Reasoning Based NLP}

\subsubsection{Predict Relevant Acts based on Case Facts}
\textbf{
\begin{itemize}
\item Description of Task
\end{itemize}
}
Predicting the relevant act of the law based on the text description of the fact is an important legal research task. This is done typically by lawyers and police while making the chargesheet. Automating this process can help layman people who don't understand law. This would help people to collect the right information about the case in a timely manner before they interact with lawyers or police. This will also increase familiarity of law in common people. 
E.g. A citizen enters text “Thieves took away my Rs. 10000 and Mobile last night....” the NLP system would return “Section 378 under Indian Penal Code”. It can also return what are the keywords in the input text description that triggered this prediction. In this case keywords could be “took away”.
\textbf{
\begin{itemize}
\item Value Proposition
\end{itemize}
}
Informed with the right section of the law, citizens can make better decisions about documents to be collected, lawyers to contact etc. This will also Increase familiarity with law among citizens
\textbf{
\begin{itemize}
\item Mapping to standard NLP Tasks
\end{itemize}
}
This maps to standard tasks of text classification.The tasks of text classification are well studied and hence this falls into the easy category.
\textbf{
\begin{itemize}
\item Similar NLP benchmarks
\end{itemize}
}
There is a similar existing benchmark about predicting which specific human rights articles and/or protocols have been violated on European Human Rights Cases \cite{chalkidis2019neural}. Another similar benchmark in Chinese language is Chinese criminal judgment prediction dataset, C-LJP \cite{xiao2018cail2018} which is a part of Chinese AI Law Challenge. In the Forum of Information Retrieval Evaluation 2019 \cite{inproceedings}, there is a dataset created which finds most relevant statutes for 50 text descriptions from 197 Sections of Acts. The description of these 197 sections is also provided.
\textbf{
\begin{itemize}
\item Dataset to be collected
\end{itemize}
}
Data needed for this benchmark are FIRs, chargesheets and judgements which describe the statements describing the incidence and applicable law sections. The labelled data could be created using an unsupervised approach like using pattern matching to extract acts and sections from text or using pretrained models for such extraction.   This data can be used to train the NLP model and evaluation. There is no human annotation needed for this benchmark.
\textbf{
\begin{itemize}
\item Evaluation Metrics
\end{itemize}
}
Evaluation of this benchmark could be done by measuring the accuracy of predicted acts and sections by comparing them with actual acts and section

\subsubsection{Statement Contradiction Discovery}
\textbf{
\begin{itemize}
\item Description of Task
\end{itemize}
}
Identification of contradictions in witnesses , accused and victims statements has a lot of impact on the verdict. These contradictions are found out by lawyers , judges and legal research teams. Automatic identification of such contradictions can significantly reduce the processing time of a case for both lawyers and judges. To achieve this, the first step is to identify the statements by multiple people about the same topic. The topic could be specific to cases like arrival time of police, observations about incidence etc. Then these statements can be compared with each other to find out potential contradictions. Example of this is as shown in table \ref{tab:contradictiondiscoverey}
\begin{table*}
\centering
\begin{tabular}{lll}
\hline
\textbf{Statement 1} & \textbf{Statement 2} &{\textbf{Prediction}} \\
\hline
Police reached  & Police reached the &  Contradiction\\ 
the site at 5pm & site at 6 pm & \\
I saw the dead body & Dead body had a lot & Entailment \\

with blood clots &  of blood marks  & \\
I saw the accused & When I reached site I saw the dead & Neutral\\
committing murder &  body with no one around & \\
\hline
\end{tabular}
\caption{Examples of Contradiction Discovery}
\label{tab:contradictiondiscoverey}
\end{table*}
These contradictions by the NLP model can be validated by humans to accept or reject them. This feedback about acceptance or rejection can be used to improve the model. 
\textbf{
\begin{itemize}
\item Value Proposition
\end{itemize}
}
This task will greatly reduce the time needed for identification of contradictions in case documents which is an important part of legal research. 
\textbf{
\begin{itemize}
\item Mapping to standard NLP Tasks
\end{itemize}
}
This maps to an NLP task called Textual Entailment also called Natural Language Inference. Depending on the dataset, these tasks can be of medium to hard complexity.
\textbf{
\begin{itemize}
\item Similar NLP benchmarks
\end{itemize}
}
Similar benchmarks focus on finding Textual Entailment in general English text (The Stanford Natural Language Inference Corpus \cite{NLI} , Recognizing Textual Entailment as part of super \cite{wang2020superglue}. Another similar benchmark \cite{COLIEE2020} is about finding a specific paragraph from case R that is relevant to a new case such that the paragraph entails decision Q of a new case.
\textbf{
\begin{itemize}
\item Dataset to be collected
\end{itemize}
}
Data needed for this benchmark would be statements recorded in FIRs and chargesheets. These then would be classified into various topics and then humans would annotate if these are contradictions or not. This data would be used to evaluate the NLP models based on accuracy of the prediction. Humans with experience in legal research would be needed to create such data. These could be lawyers or assistants who perform legal research. A custom annotator tool needs to be built to allow such human annotations. This tool would show the suggested contradictions .The answer suggested can be accepted by the expert or he can change it to an appropriate one. Experts will also add the contradictions that are not present in suggestions. This suggestion would reduce the human processing time significantly.
\textbf{
\begin{itemize}
\item Evaluation Metrics
\end{itemize}
}
This benchmark can be evaluated by comparing the human labeled triplets with model predictions. Accuracy of such predictions can be used as a metric.

\subsubsection{Sentence Rhetorical Roles Prediction}
\textbf{
\begin{itemize}
\item Description of Task
\end{itemize}
}
Although the style of writing a judgement varies by the Judge, most of the judgements have an inherent structure. Giving structure to the judgment text is important for many Information retrieval and other downstream tasks. Sentence Rhetorical Roles prediction means identifying what role a sentence is playing in the judgement. Recent research has created a small dataset and built an ML model to predict the rhetorical role of each sentence.   
\textbf{
\begin{itemize}
\item Value Proposition
\end{itemize}
}
The identification of right section of the judgement narrows the text to focus for a given task. E.g. If a person wants to know the final decision then it could be found in the section marked as “Current Court Decision”. If Someone wants to know the description of the case then it could be found in “Facts” section.  
The rhetorical roles identification would also help significantly in creating summary of the judgements and semantic search. 
\textbf{
\begin{itemize}
\item Mapping to standard NLP Tasks
\end{itemize}
}
This would fit into the task of multi-class text classification where each sentence is assigned with a rhetorical role. The rhetorical role of a sentence is also dependent on the previous and next sentences.
\textbf{
\begin{itemize}
\item Similar NLP benchmarks
\end{itemize}
}
There are some datasets released about this task but there is no benchmark. The dataset published by \cite{bhattacharya2019identification} is very small and noisy in nature.
\textbf{
\begin{itemize}
\item Dataset to be collected
\end{itemize}
}
The manual annotations at the sentence level about which sentence belongs to what rhetorical role need to be collected. The size of the dataset should not be as small as the one mentioned in the original paper by \cite{bhattacharya2019identification}. 
\textbf{
\begin{itemize}
\item Evaluation Metrics
\end{itemize}
}
The accuracy of the prediction could be used as evaluation metric

\section{Useful NLP components across Benchmarks}
This section talks about some building blocks which are useful across multiple benchmarks.
\subsection{Indian Legal BERT}
BERT is a pre-trained language model that helps to use intelligence from vast amounts of unlabelled text  and use it for doing specific tasks where less data is available. Many of the current BERT models are trained on General English Text like Wikipedia. Creating BERT specifically in the context of Indian Law will help many of the Indian Legal Tasks. This model will learn the vocabulary of the Indian Legal system and semantics of it. Creating such domain specific BERT models has shown promising results in literature. This BERT model will help across the multiple Indian Legal NLP benchmarks.
\subsection{Legal Entities Extractor}
Extraction of legal entities like court name, judge name,  parties involved, acts and sections etc. are useful. These would act as inputs to multiple benchmarks. Having a component which understands the language used by Indian Law would make it perform better than other out of the box entity recognition components.
\subsection{Cleaned Textual Repository}
To overcome the challenges about public datasets mentioned in the previous section, it is important to create an open textual repository of indian legal text. Having such a repository would save a lot of time in scraping the legal websites and bringing them to unified format. This will also provide data for human annotations for multiple benchmarks. 
\begin{figure*}[htp]
    \centering
    \includegraphics[scale=0.6]{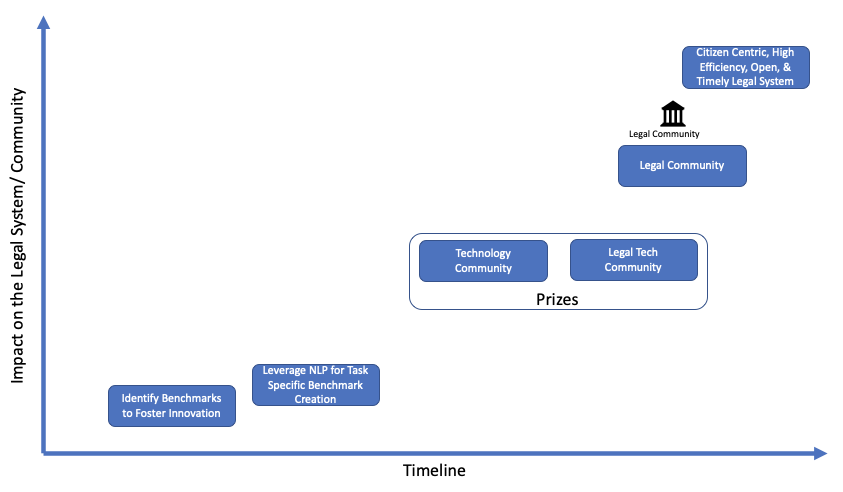}
    \caption{Operationalizing NLP Benchmarks}
    \label{fig:operationalization}
\end{figure*}

\section{Operationalizing Indian NLP Benchmarks}

The timeline chart below shows how the NLP benchmark will start showing benefits for the Legal system with time. Starting point is identification of NLP benchmarks and doing efforts and impacts study of these NLP benchmarks. This white paper is part of this step. In the next step, relevant datasets would be collected (either using humans or automatically from the text based on the benchmark). Baseline NLP models would be created which indicate bare minimum evaluation metric values. The datasets would be split into two sets: training data and testing data. Training data would be provided to the participants of the competition. Testing data would be kept hidden and would be used for evaluation purposes. 

The competitions would be launched with these datasets along with prizes for winners. Many times the research value of such solutions is more important than the prize money. So it is important to publish the dataset in reputed journals which attract the ML community. For legal communities like startups \& legal product development companies, the benchmark would present opportunities to objectively show how good their product is.
Once the Indian Legal NLP benchmarks become widely recognized in the ML community and legal community then the benefits of such benchmarks start to show up. Many of the researchers open source the code they developed for research for the benchmarks. This helps the overall community and researchers can spend their time in solving more complex problems. Since these benchmarks present challenging tasks they will open up more possibilities and applications. All these things would help Indian Legal systems to be more efficient, open and citizen centric.

\section*{Acknowledgements}

This paper is funded by EkStep Foundation

\bibliography{anthology,custom}
\bibliographystyle{acl_natbib}

\appendix

\end{document}